# Self-Supervised Surgical Instrument 3D Reconstruction from a Single Camera Image

Ange Lou[1], Xing Yao[2], Ziteng Liu[2], Jintong Han[2], Jack Noble[1]
Department of Electrical Engineering[1], Vanderbilt University, Nashville TN, USA
Department of Computer Science[2], Vanderbilt University, Nashville TN, USA
{ange.lou, xing.yao, ziteng.liu, jintong.han, jack.noble}@vanderbilt.edu

**Abstract**

Surgical instrument tracking is an active research area that can provide surgeons feedback about the location of their tools relative to anatomy. Recent tracking methods are mainly divided into two parts: segmentation and object detection. However, both can only predict 2D information, which is limiting for application to real-world surgery. An accurate 3D surgical instrument model is a prerequisite for precise predictions of the pose and depth of the instrument. Recent single-view 3D reconstruction methods are only used in natural object reconstruction and do not achieve satisfying reconstruction accuracy without 3D attribute-level supervision. Further, those methods are not suitable for the surgical instruments because of their elongated shapes. In this paper, we firstly propose an end-to-end surgical instrument reconstruction system — **S**elf-supervised **S**urgical **I**nstrument **R**econstruction (**SSIR**). With SSIR, we propose a multi-cycle-consistency strategy to help capture the texture information from a slim instrument while only requiring a binary instrument label map. Experiments demonstrate that our approach improves the reconstruction quality of surgical instruments compared to other self-supervised methods and achieves promising results.

## INTRODUCTION

Recent single-view 3D object reconstruction methods are widely applied to recover the shape and texture information of nature objects [1]. It is a long-standing problem in computer vision with various applications like 3D scene analysis, robot navigation, and virtual/augmented reality. Previous methods like 3DMM [2] and SMPL [3] usually fit the parameters of a 3D prior morphable model, which is too time-consuming and expensive to various objects.

Deep learning achieves promising performance in various computer vision areas, such as segmentation [4, 5], classification [6] and object detection [7]. It has also been widely used in supervised 3D object reconstruction which directly minimize the difference between ground truth 3D annotations and predictions. However, compared with 2D annotations, 3D annotations are more difficult to obtain, thus 2D supervised reconstruction methods become a more popular topic. The key part of 2D supervised reconstruction is Differential Rendering (DR) [8] which allowed the gradients of 3D objects to be calculated and propagated through 2D images.

In the image-guided surgery area, accurate surgical instrument tracking techniques can help surgeons track and localize the position of surgical tools. Most recent tracking techniques are based on 2D image segmentation [9, 10, 11] and object detection [12], which have very limited information to estimate instruments' pose and position in 3D space. An accurate 3D instrument model is a prerequisite for estimating its motion (pose and depth changes) in a surgical video and for providing more comprehensive feedback to surgeons. However, the elongated shape surgical instrument is difficult to be reconstructed by any current existing 2D supervised method and the texture information of surgical tools is more challenging by learning from networks.

In traditional cochlear implant (CI) surgery, an electrode array is inserted into the cochlea relying on the surgeon's expertise to determine the insertion vector and overall insertion depth, even though the surgeon is only able to visualize the cochlear entry point blind to the intra-cochlear anatomy and intra-cochlear insertion path. A system that could provide the localization and pose information for the electrode insertion tool from the surgical microscope images could be used to ensure the insertion vector and overall depth match an optimized pre-operative plan based on CT imaging [13, 14] to enable correcting the insertion trajectory in the operating room. In this work, we aim to develop such a system to reconstruct electrode insertion instruments and to recover shape and texture information using only single frame microscopic images. We call our approach **S**elf-supervised **S**urgical **I**nstrument **R**econstruction (**SSIR**). Though we test it here on CI insertion tools, the method is generally applicable to many other types of surgical tools. A summary of our contributions is:

(1) We propose SSIR to reconstruct a 3D surface mesh of arbitrary objects from a collection of surgical video frames. It is an end-to-end training system that relies only on a silhouette mask annotation of the object of interest.
(2) We propose a multi-cycle-consistency learning strategy to guarantee texture information of surgical instruments can be recovered for slim instruments.
(3) Our SSIR method outperforms the current state-of-the-art mesh reconstruction method [15] on our CI dataset.

## METHOD

For a given collection of CI insertion instrument images, we used a semi-supervised segmentation state-of-the-art—Min-Max Similarity (MMS) [16] to obtain their silhouette mask, we aim to train encoders to predict 3D reconstruction attributes — camera, shape, texture, and light parameters.

### A. Differentiable Rendering

The 3D mesh model can be represented as $O(S,T)$. $S \in \mathbb{R}^{V \times 3}$ represents the mesh vertices, and $V$ is the total number of vertices. $T \in \mathbb{R}^{\mathcal{H} \times \mathcal{W} \times 3}$ represents the texture map with resolution $\mathcal{H} \times \mathcal{W}$, aka the "UV" map. We defined $C = (a, e, d)$ as parameters of the rendering camera, in which $a \in [0°, 360°]$, $e \in [-90°, +90°]$, and $d \in (0, +\infty]$ stand for the azimuth, elevation and distance parameters. Light attribute $L \in \mathbb{R}^l$ is modeled by Spherical Harmonics [17] and $l$ is the dimension of coefficient. For a given 3D attribute $A = [C, L, S, T]$, a 3D object $O(S,T)$ can be rendered as the 2D image and its binary instrument label mask $X^r = [I^r, M^r]$ under the fixed camera view $C$ and lighting condition $L$. $X^r$ is the concatenation of the rendered RGB image $I^r \in \mathbb{R}^{H \times W \times 3}$ and silhouette mask $M^r \in \mathbb{R}^{H \times W \times 1}$, $H \times W$ represents the resolution of input images and silhouette mask. And the rendering process can be represented as equation (1):

$$X^r = R(A) = R([C, L, S, T]) \tag{1}$$

where $R$ is a differentiable renderer that does not contain any trainable parameters.

### B. Reconstruction Network

The reconstruction network is a combination of four sub-encoders which are used to predict the 3D attributes $A$ and the architecture is shown in figure 1. The $i_{th}$ input $X_i = [I_i, M_i]$ ($i = 1, 2 \dots, N$ and $N$ is the number of training samples) is sent to the reconstruction model $E_\theta$ to predict 3D attributes $A_i$ as

$$A_i = E_\theta(X_i) = [C_i, L_i, S_i, T_i] \tag{2}$$

Where $\theta$ is the parameters of encoders. $E_\theta(\cdot) = \{E_c, E_l, E_s, E_t\}$ where $E_c, E_l, E_s, E_t$ represents encoders of camera, light, shape and texture respectively.

The camera encoder $E_c$ predicts a 4D vector $[a_x, a_y, e, d]$, where $e$ and $d$ represent elevation and distance of parameters. And $(a_x, a_y)$ denotes the Cartesian coordinates of azimuth, and $a = atan2(a_x, a_y)$. The shape encoder $E_s$ predicts the relative shape change $\Delta S \in \mathbb{R}^{V \times 3}$ to an initial spherical mesh $S_0 \in \mathbb{R}^{V \times 3}$, and the object shape can be represented as $S = \Delta S + S_0$. Similar to CMR [18], the texture encoder $E_t$ predicts a 2D flow map and then applies spatial transformation [19] to generate a texture UV map. The light encoder $E_l$ predicts an $l$-dimension vector as the Spherical Harmonics model coefficients.

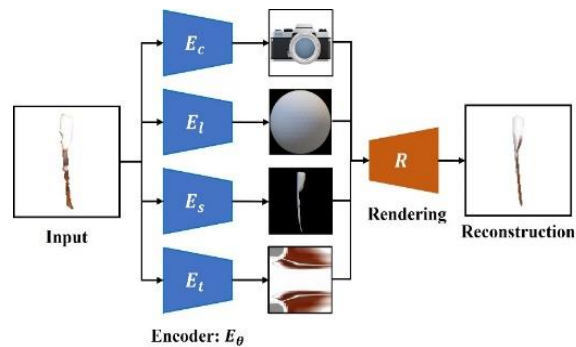

Figure 1. Reconstruction model.

### C. Self-supervised Reconstruction using multi-cycle consistency

In this section, we will summarize how SSIR trains the encoder to predict 3D attributes under 2D level supervision and 3D level self-supervision. Unlike the natural object reconstruction like CUB-200-2011 dataset [20], the texture of the CI insertion instrument is more difficult to detect due to its elongated shape. Thus, we propose multi-cycle-consistency to help the reconstruction model capture a texture map from the input image. The complete network we propose to train the reconstruction model using multi-cycle consistency is shown in figure 2. As seen in the figure, two independent reconstruction models $E_{\theta\_1}$ and $E_{\theta\_2}$ are trained simultaneously to predict two 3D attributes $A_{i\_1}$ and $A_{i\_2}$. On the 2D supervision routes on the left side, these attributes are rendered by the differentiable renderer to generate rendered data and supervised by $\mathcal{L}_{2D}$ (defined below). On another route on the right, we generate new views by mixing 3D attributes $\{A_{i\_1}, A_{i\_2}\}$ with 3D attributes $\{A_{j\_1}, A_{j\_2}\}$ corresponding to images randomly sampled from a different training batch. The mixture forms a new set of 3D attributes for a new view $A_{ij}$ representing an interpolation of the 4 input attribute sets:

$$A_{ij} = 0.5 \cdot [(1-\alpha_1) \cdot A_{i\_1} + \alpha_1 \cdot A_{j\_1}] + 0.5 \cdot [(1-\alpha_2) \cdot A_{i\_2} + \alpha_2 \cdot A_{j\_2}] \quad (3)$$

where $\alpha_1$ and $\alpha_2$ are sampled from a uniform distribution $U \sim (0,1)$. Figure 3 further illustrates the concept of multi-cycle-consistency. 2D images $X_i$ and $X_j$ should be cycle consistent with their 3D attributes, and the 3D attributes synthesized using equation (8) should be cycle consistent with their corresponding synthetic 2D images. This augmentation approach not only enhances the four sub-encoders of the reconstruction network (see Figure 1) but also improve the quality of rendered images.

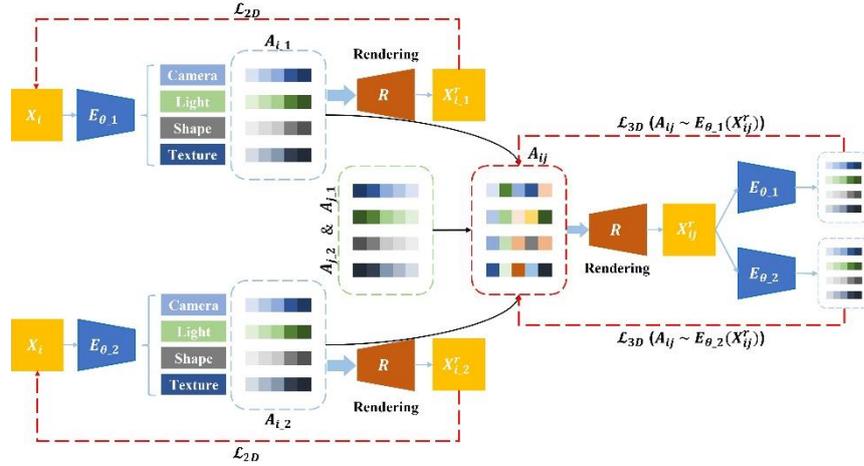

Figure 2. Architecture of SSIR

**2D level supervision.** The differentiable renderer $R$ can render specified 2D images $X_{i,\theta}^r = R(E_\theta(X_i))$ with 3D attributes generated by a given reconstruction model, $E_\theta$. For 2D level supervision we apply two terms. First, we apply $L_1$ distance to measure the distance between the foreground of rendered and input images as

$$\mathcal{L}_{\text{img}}^\theta = \frac{1}{N} \sum_{i=1}^{N} \left\| I_i \odot M_i - I_{i,\theta}^r \odot M_{i,\theta}^r \right\|_1, \quad (4)$$

where $\odot$ denotes element multiplication and $\theta$ corresponds to reconstruction network $E_{\theta\_1}$ or $E_{\theta\_2}$. We also apply mask IoU to measure the similarity between input and rendered silhouette mask,

$$\mathcal{L}_{\text{sil}}^\theta = \frac{1}{N} \sum_{i=1}^{N} \left( 1 - \frac{\left\| M_i \odot M_{i,\theta}^r \right\|_1}{\left\| M_i + M_{i,\theta}^r - M_i \odot M_{i,\theta}^r \right\|_1} \right). \quad (5)$$

The overall 2D level supervision loss can be represented as the sum of the image and silhouette mask loss,

$$\mathcal{L}_{2D}^\theta = \lambda_{img} \mathcal{L}_{\text{img}}^\theta + \lambda_{sil} \mathcal{L}_{\text{sil}}^\theta. \quad (6)$$

**3D level self-supervision.** The 2D level supervision can only optimize the reconstruction model under the original view. To satisfy the requirements for reconstructing the 3D mesh from a different view, we apply the interpolation method in Eqn (1) discussed above to generate a new view from new 3D attributes and use multi-cycle consistency to perform self-supervised training. The self-supervised loss can be formulated as

$$\mathcal{L}_{3D}^\theta = \frac{1}{N} \sum_{i=1}^{N} \left\| E_\theta\left(R(E_\theta(X_i))\right) - E_\theta(X_i) \right\|_1. \quad (7)$$

The encoder $E_\theta$ predicts 3D attributes $E_\theta(X_i)$. Then a new rendered image $R(E_\theta(X_i))$ is reconstructed from the attributes. The rendered image is sent again to encoder $E_\theta$ to receive a new 3D attributes prediction $E_\theta\left(R(E_\theta(X_i))\right)$, and we aim to minimize the loss between the final predicted attributes $E_\theta\left(R(E_\theta(X_i))\right)$ and the initial ones $E_\theta(X_i)$. For the original input data, which already has 2D level supervision, we do not need to apply 3D self-supervision. But for the new view generated attributes $A_{ij}$, which do not have a 2D level label, it is necessary to apply self-supervision method to update the reconstruction model.

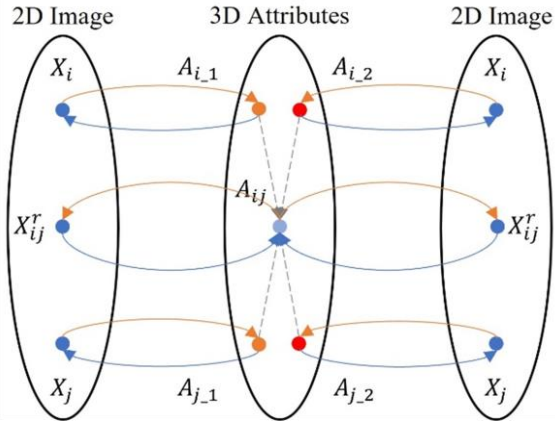
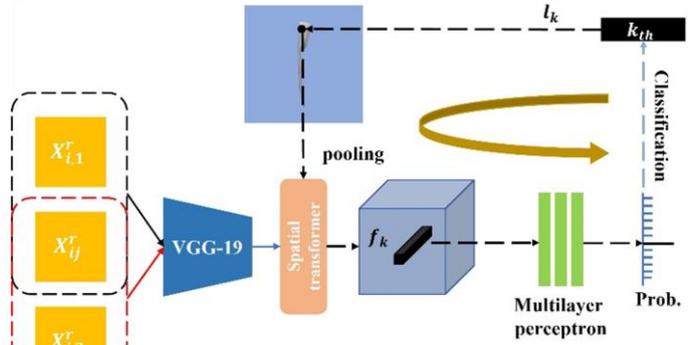

Figure 3. Multi cycle consistency    Figure 4. Landmark consistency

**Landmark Consistency.** To further improve the quality of reconstructed images, we applied landmark consistency between the two reconstruction models, which is another self-supervised method. As shown in figure 4, we use a pre-trained VGG-19 [21] to extract a feature map $F_{i,\theta}$ from the concatenation of $X_{i,\theta}^r = R(E_\theta(X_i))$ and $X_{ij}^r$. The coordinates of the vertices in the 3D shape attribute mesh can be represented as $\{l_k\}_{k=1}^V$. In the spatial transformer, $l_k$ is projected to 2D image space as a landmark and pooled with the local feature $f_k^\theta$ extracted from $F_{i,\theta}$, where $\theta$ can be $\theta\_1$ or $\theta\_2$ corresponding to the two reconstruction models, by spatial transformation [19] from the feature maps. Finally, we apply a multilayer perceptron $D_\emptyset(\cdot)$ with weight $\emptyset$ to predict index of each landmark. Since the classification label of $f_k^\theta$ is $k$, the self-supervised classification loss function can be represented as

$$\mathcal{L}_{\text{LC}}^\theta = -\frac{1}{N}\sum_{i=1}^N \sum_{k=1}^V v_k^\theta y_k \log\left(D_\emptyset(f_k^\theta)\right) \tag{8}$$

where $y_k$ is a one-hot vector with size $V$, in which only $k_{th}$ value is 1, and $v_k^\theta$ indicates if the $k_{th}$ vertex is visible in $X_{i,\theta}^r$. This self-supervision term leads to two reconstruction models that have similar performance in landmark classification using the co-trained multi-layer perceptron. Using features extracted from $\{X_{i,\theta}^r, X_{ij}^r\}$ ensures corresponding landmarks are classified at the same spatial location between $X_{i,\theta\_1}^r$ and $X_{i,\theta\_2}^r$.

**Overall Loss.** The final system contains two encoders (dual view), and the overall loss for training $E_{\theta\_1}$ and $E_{\theta\_2}$ can be represented as $\mathcal{L} = 0.5 \cdot (\lambda_{2D}\mathcal{L}_{2D}^{\theta\_1} + \lambda_{3D}\mathcal{L}_{3D}^{\theta\_1} + \lambda_{LC}\mathcal{L}_{LC}^{\theta\_1}) + 0.5 \cdot (\lambda_{2D}\mathcal{L}_{2D}^{\theta\_2} + \lambda_{3D}\mathcal{L}_{3D}^{\theta\_2} + \lambda_{LC}\mathcal{L}_{LC}^{\theta\_2})$.

**Dataset.** We selected four CIs insertion videos that contain a complete view of the inserting instrument. To get the segmentation mask, we first labelled 20 image frames for each video and then used those to train and apply our previously proposed Min-Max Similarity algorithm [16] to predict segmentation masks for the rest of the image frames. The image and segmentation samples as shown in Figure 5.

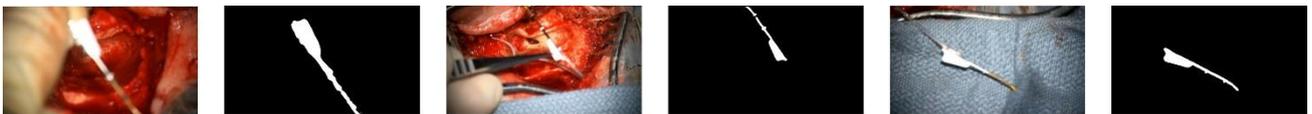

Figure 5. Images and segmentation masks for CIs inserting instrument.

**Results**

To measure the 3D reconstruction accuracy, we use mask IoU to evaluate the similarity between the rendered and original data. For the new-view reconstruction, we apply an image generation metric, Frechet Inception Distance (FID), to evaluate performance. We calculate the mean FID of the synthesized images in the novel view from $0°$ to $360°$ at an interval of $30°$. The results are shown in Table 1 and Figure 6.

Table 1. Comparison between the SMR and our SSIR

|  | Mask IoU | Reconstruction FID ↓ | Rotation FID ↓ |
|---|---|---|---|
| **SMR** | 0.868 | 180.3 | 211.4 |
| **SSIR (ours)** | 0.867 | **121.4** | **126.6** |

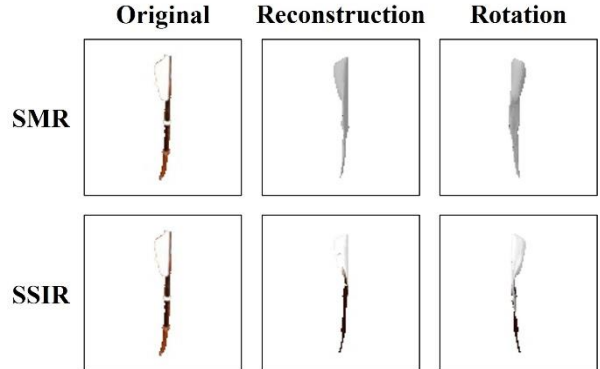

Figure 6. Reconstruction results.

## DISGUSSIONS

We propose a self-supervised surgical instrument reconstruction (SSIR) with the multi-cycle-consistency loss to reconstruct CIs inserting instruments. Compared with existing methods like CMR [22], CSM [22], DIB-R [23] and UMR [24], SSIR does not require camera parameters, instrument templates, predefined landmarks, or subregion labels [25]; but can capture accurate shape and texture information from both original and generated novel views as shown in Table 1 and Figure 6.

## CONCLUSIONS

SSIR is the first end-to-end self-supervised 3D reconstruction model which can be used to reconstruct surgical tools from single 2D frames. Compared with the state-of-the-art SMR, our SSIR successfully recovers texture information even for difficult, highly elongated shapes. Ongoing work to be presented at the conference includes measuring the 3D reconstruction shape error when compared to a ground truth instrument model, evaluating the use of this method for coordinate localization of the instrument in the operating room, a comprehensive ablation study on our architecture and hyperparameters, and testing the method on other surgical instruments.

878).